\documentclass[10pt,twocolumn,letterpaper]{article}

\usepackage{iccv}              

\definecolor{iccvblue}{rgb}{0.21,0.49,0.74}
\usepackage[pagebackref,breaklinks,colorlinks,allcolors=iccvblue]{hyperref}
\usepackage[accsupp]{axessibility}
\usepackage{stfloats} 
\usepackage{float} 
\usepackage{multicol}
\usepackage{multirow}
\usepackage{kotex}
\usepackage{booktabs}
\usepackage{algorithm}
\usepackage{algpseudocode}
\usepackage[dvipsnames]{xcolor}
\usepackage{tikz}
\usepackage{arydshln} 
\usepackage{subcaption}
\usepackage{afterpage}
\usepackage[utf8]{inputenc}
\def\paperID{6054} 
\def\confName{ICCV}
\def\confYear{2025}

\def\Ours{ConceptSplit}

\title{\Ours: Decoupled Multi-Concept Personalization of Diffusion Models \\ via Token-wise Adaptation and Attention Disentanglement}

\author{
    Habin Lim\textsuperscript{1} \qquad
    Yeongseob Won\textsuperscript{2} \qquad
    Juwon Seo\textsuperscript{2} \qquad
    Gyeong-Moon Park\textsuperscript{1}\thanks{Corresponding author} \\\\
    \textsuperscript{1}Korea University \qquad\qquad\qquad\qquad
    \textsuperscript{2}Kyung Hee University\\
    {\tt\small \{ha001211, gm-park\}@korea.ac.kr}, \qquad
    {\tt\small \{wysgene19, jwseo001\}@khu.ac.kr}
}

\begin{document}

\maketitle
\begin{abstract}
In recent years, multi-concept personalization for text-to-image (T2I) diffusion models to represent several subjects in an image has gained much more attention. The main challenge of this task is “concept mixing”, where multiple learned concepts interfere or blend undesirably in the output image.
To address this issue, in this paper, we present \Ours, a novel framework to split the individual concepts through training and inference. Our framework comprises two key components. First, we introduce Token-wise Value Adaptation (ToVA), a merging-free training method that focuses exclusively on adapting the value projection in cross-attention. Based on our empirical analysis, we found that modifying the key projection,  a common approach in existing methods, can disrupt the attention mechanism and lead to concept mixing. Second, we propose Latent Optimization for Disentangled Attention (LODA), which alleviates attention entanglement during inference by optimizing the input latent. Through extensive qualitative and quantitative experiments, we demonstrate that \Ours achieves robust multi-concept personalization, mitigating unintended concept interference.
Code is available at \href{https://github.com/KU-VGI/ConceptSplit}{https://github.com/KU-VGI/ConceptSplit}.
\end{abstract}    
\section{Introduction}
\label{intro}

\begin{figure*}[t]
    \centering
    \includegraphics[width=\linewidth]{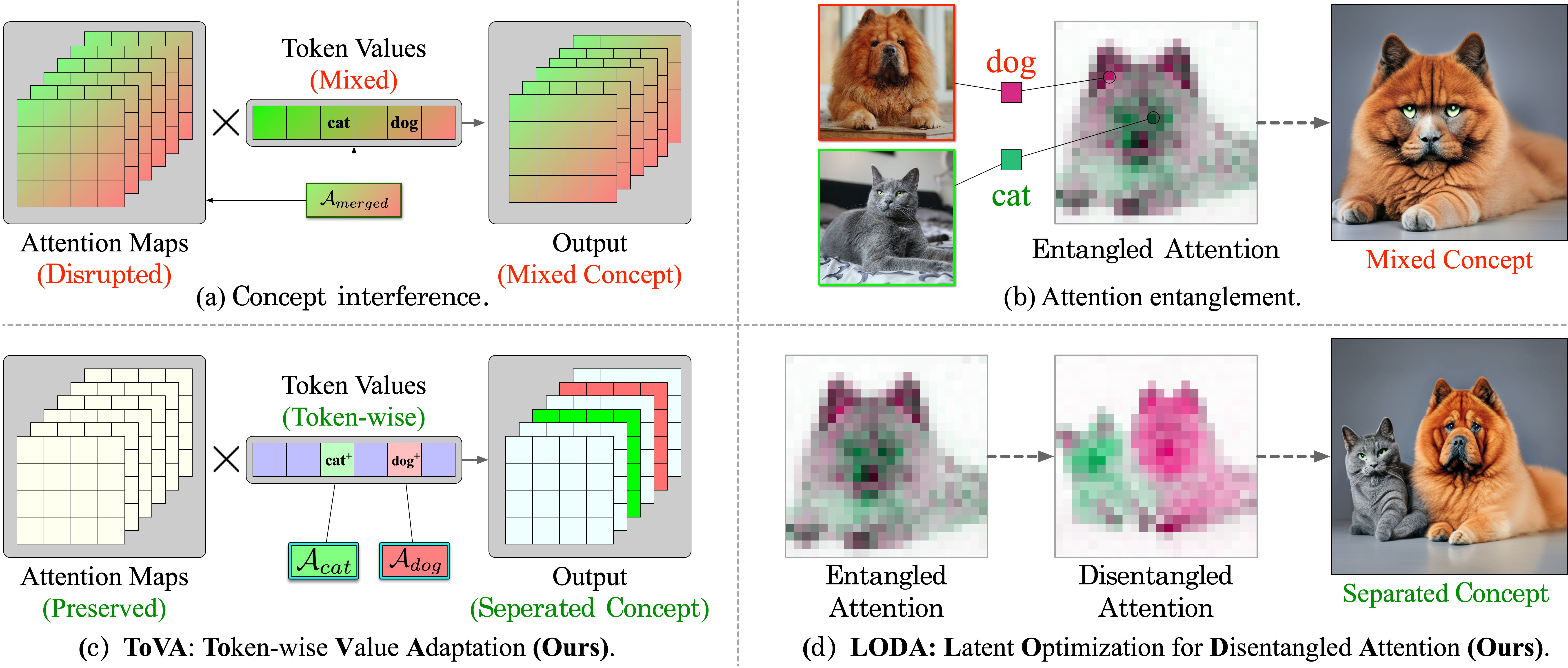}
     \caption{{The goals of our framework \textbf{\Ours} are twofold. (1) Preventing concept interference in the adapter approach (denoted as $\mathcal{A}$), while preserving the token-attention binding capacity of T2I models. (2) Separating entangled attention, which would otherwise result in a mixed representation of learned concepts.}}
\label{fig:teaser}
\end{figure*}

With recent significant advancements in Text-to-Image (T2I) diffusion models, such as Stable Diffusion \cite{rombach2022high, podell2024sdxl, esser2024scalingrectifiedflowtransformers}, personalization has gained increasing attention, becoming the mainstream approach for customized T2I models.

Traditional personalization techniques primarily focus on general-purpose methods for single-concept personalization, such as DreamBooth \cite{ruiz2023dreambooth} and Textual Inversion \cite{gal2022textualinversion}. However, real-world applications often require the simultaneous integration of multiple concepts, such as several custom subjects interacting with a unique background. This necessity has led to the emergence of multi-concept personalization, which aims to enable T2I models to coherently blend multiple personalized subjects while maintaining high fidelity and consistency.

Recently, numerous methods have been proposed to tackle this appealing task, aiming to ensure robust and disentangled multi-concept personalization \cite{kumari2023custom, liu2023cones, liu2023cones2, smith2024continuallora, james2024stamina, sun2024cyw, gu2024mixofshow, kong2024omg, zhu2024multiboothgeneratingconceptsimage}.
However, existing studies face a critical challenge in handling multi-concept personalization, a concept mixing problem \cite{kumari2023custom}, where multiple learned concepts interfere with or influence unintended ones. For example, as shown in Figure \ref{fig:teaser} (b), the learned concept of a dog can influence the concept of a cat during the personalization, causing the generated image of the single dog to resemble features of the cat, \ie, two concepts are undesirably mixed.

To deal with this issue, there are two main approaches: (1) adapter-based approaches to learn each concept independently \cite{gu2024mixofshow, kong2024omg, smith2024continuallora, sun2024cyw, james2024stamina} and (2) text embedding-based approaches to selectively modify text embeddings of desired tokens \cite{gal2022textualinversion, vinker2023concept, liu2023cones2, voynov2023p+, pang2023cross, zhang2023compositional}.

However, existing adapter-based approaches typically require a merging of adapters at inference time, which causes interference between learned concepts and ultimately leads to concept mixing (Figure \ref{fig:teaser} (a)). Moreover, we argue that existing text embedding-based approaches interfere with the token-attention binding capacity, the ability to correctly map each token to its corresponding latent region \cite{hertz2023prompttoprompt,Liu2024sdunderstand}, resulting in an attention disruption problem. We further analyze this problem in Figure \ref{fig:entropy} and \ref{fig:disruption}.

Based on the limitations of existing works, we present three major issues that should be considered to perform multi-concept personalization effectively. (1) \textbf{Merging-free personalization}: when learning a new concept, the training strategy needs to be designed in a merging-free manner to avoid unwanted influence during inference.
(2) \textbf{Preserving token-attention binding capacity}: since personalization often proceeds on very few images, leveraging the pre-trained model's token-attention binding capacity is crucial for efficiently learning new concepts without mixing. (3) \textbf{Mitigating attention entanglement}: during inference, concepts may unintentionally influence each other due to the T2I model’s token-attention mechanism, especially for semantically similar concepts. To ensure accurate personalization, disentangling attention while maintaining the token-attention binding capacity is necessary.

In this paper, to address the above issues, we propose a novel framework, \textbf{\Ours}, designed to selectively learn and disentangle concepts. Our framework comprises two key components. The first component, \textbf{To}ken-wise \textbf{V}alue \textbf{A}daptation (\textbf{ToVA}), is a novel adaptation approach designed to prevent concept mixing. Instead of training every component of the attention mechanism, ToVA exclusively trains adapters for the value projection of the desired tokens. To maintain the token-attention binding capacity, ToVA does not modify the key in the cross-attention mechanism, thereby preventing disruption. At inference time, ToVA attaches the learned adapters to corresponding tokens to update their values only without merging.
The second component, \textbf{L}atent \textbf{O}ptimization for \textbf{D}isentangled \textbf{A}ttention (\textbf{LODA}), mitigates attention entanglement by optimizing the input latent, instead of modifying the attention map to preserve the token-attention binding capacity while achieving attention disentanglement. This approach allows the model to naturally produce the disentangled attention in a self-guided manner. The extensive experiments demonstrate that our \textit{merging-free} personalization method while \textit{preserving token-attention binding capacity} with \textit{attention disentanglement} successfully performs multi-concept personalization without concept mixing, showcasing state-of-the-art performances on benchmark scenarios.

Our main contributions can be summarized as follows:
\begin{itemize}
    \item We propose a novel framework \textbf{\Ours}, which shows robust multi-concept personalization without concept mixing. Our method comprises \textbf{ToVA}, a novel training method for merging-free multi-concept personalization, and \textbf{LODA}, a self-guided approach that optimizes latent vector to enhance attention disentanglement among concepts.
    \item We analyze the attention disruption problem in text embedding-based methods, which is our key motivation for the proposed method. In addition, we present the analysis of attention entanglement in concept mixing and suggest an effective solution supported by intuitive visualizations.
    \item Through comprehensive quantitative and qualitative experiments, we demonstrate that our method offers superior personalization capability compared to existing multi-concept personalization techniques.
\end{itemize}\

\section{Related Work}
\label{sec:method}
\begin{figure*}[t]
    \centering
    \includegraphics[width=\linewidth]{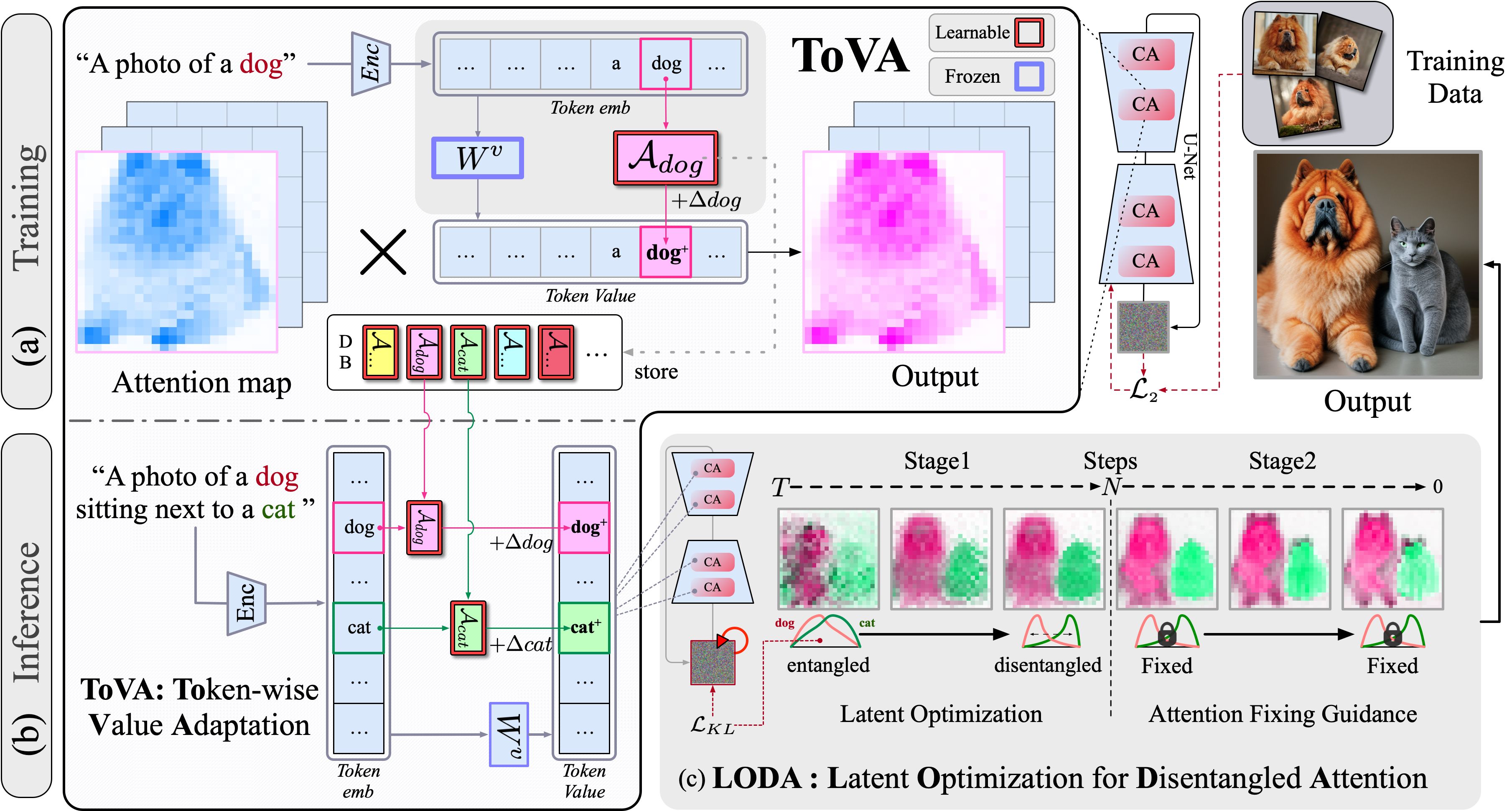}
    \caption{\textbf{Overview of our framework \Ours.}
(a) \textit{Training Phase}: Adapters are applied exclusively to target tokens, modifying only their values while preserving the model’s baseline attention capacity. These adapters are then stored in a database (DB).
(b) \textit{Inference Phase}: Pre-trained adapters are dynamically attached to desired tokens, enabling selective value modulation without weight merging. This ensures minimal interference with unrelated concepts. (c) \textit{LODA}: We further optimize the latent space for 
$N$ inference steps to disentangle attention (stage 1), then fixing the disentangled attention to enable natural generation of distinct objects without continuous optimization (stage 2).}
    \label{fig:mainfigure}
\end{figure*}

\paragraph{Multi-concept Personalization.}
Multi-concept personalization aims to generate images that simultaneously reflect multiple user-specified concepts. Custom Diffusion \cite{kumari2023custom} pioneered extending this task to a multi-concept scenario by composing multiple customized concepts in a single image while highlighting the limitations of personalization when dealing with similar concepts. Subsequently, numerous works have sought to address these challenges, employing methods such as fine-tuning \cite{liu2023cones, han2023svdiff}, textual embeddings \cite{liu2023cones2}, and adapter techniques \cite{sun2024cyw, james2024stamina, gu2024mixofshow, Hyung2024magicapture, zhu2024multiboothgeneratingconceptsimage}. Despite these promising advances, preserving the uniqueness of each concept remains challenging. This paper further examines three key issues—merging in adapters, attention disruption, and entangled attention—to achieve more robust multi-concept personalization.

\paragraph{Adapter-based Finetuning.}
Adapter-based methods, popularized in NLP by LoRA \cite{hu2021lora} and its variants \cite{liu2024dora, Seo2023left, zhao2024galore, Agiza2024mtlora, tian2024hydralora}, have recently gained traction in computer vision for efficient parameter adaptation. These approaches enable robust few-shot domain adaptation while preserving pre-trained model integrity. In parallel, personalized fine-tuning leverages adapters to encode concepts into pre-trained models, either for single-subject customization (e.g., identity preservation in \cite{shi2023instantbooth, gal2023designing, wang2024instantid}) or multi-concept synthesis (e.g., \cite{sun2024cyw, gu2024mixofshow, zhu2024multiboothgeneratingconceptsimage}). However, prior work predominantly focuses on optimizing adapter usage during inference (e.g., via attention masking or gradient-based fusion). We instead propose novel training strategies that address interference between adapters, a critical limitation in multi-concept scenarios.
\paragraph{Image Synthesis via Attention Control.}
Recent advances in image synthesis have demonstrated the power of attention control to enhance concept representations \cite{phung2024ground, hertz2023prompttoprompt, chefer2023attendandexcite, wang2023compositionaltexttoimagesynthesisattention}. For instance, Prompt-to-Prompt \cite{hertz2023prompttoprompt} was the first to show that cross-attention maps in Stable Diffusion \cite{rombach2022high} capture rich semantic relations between tokens and latent vectors, enabling effective prompt-based image editing. Likewise, Attend-and-Excite (AaE) \cite{chefer2023attendandexcite} strengthens attention map activations for desired tokens, ensuring that key elements stand out. Building on these insights, our work also leverages attention control—but with a distinct twist. We optimize the latent space so that the attention distributions for each concept diverge and avoid overlap, thereby reducing interference among concepts.

\section{ConceptSplit}
In this section, we propose a novel framework, coined \Ours, which comprises two stages: training and inference.
First, in Section \ref{sec:tova}, we analyze the problem of merging adapters and introduce the motivation behind our training method, ToVA, followed by Section \ref{sec:attentiondisruption} with detailed explanation of our approach.
Second, in Section \ref{sec:LODA}, we address the issue of attention entanglement during inference and demonstrate how our proposed LODA effectively resolves this problem. Figure \ref{fig:mainfigure} illustrates our framework.

\subsection{Token-wise Value Adaptation}
\label{sec:tova}
While typical adapter-based approaches learn each concept independently, they require a merging of adapters at inference time, which causes interference among learned concepts and ultimately leads to concept mixing. To address this issue, we propose a new merging-free adaptation technique, called Token-wise Value Adaptation (ToVA), which considers multi-concept inference while mitigating attention disruption issues in text embedding-based methods.

\begin{figure*}[t]
  \centering
  \begin{tabular}{c: c}
    \begin{minipage}{0.26\linewidth}
      \centering
      \includegraphics[width=\linewidth]{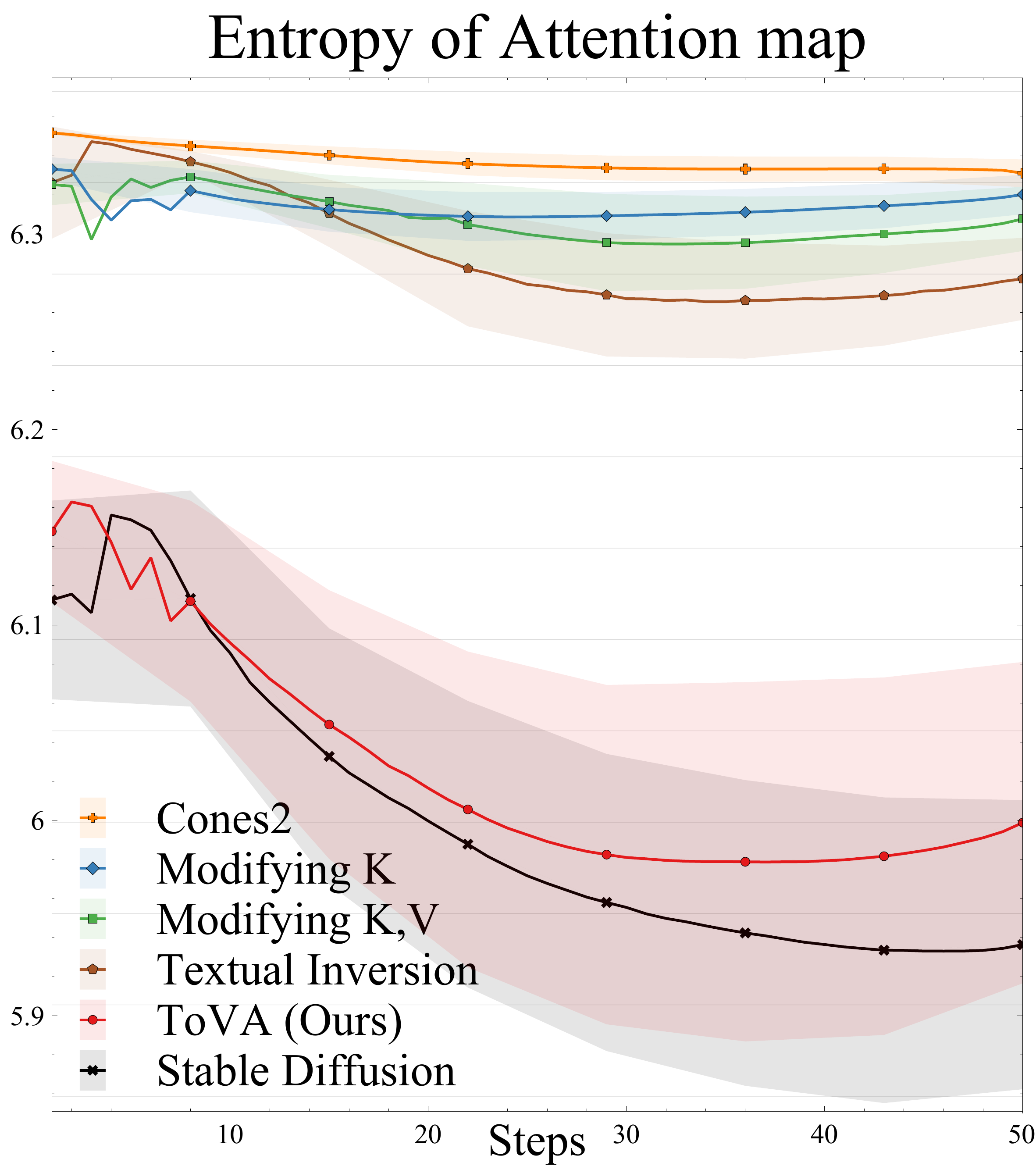}
        \caption{
        \textbf{Entropy variation in attention maps.} See more details in Supplementary B.
        }
      \label{fig:entropy}
    \end{minipage}
    &
    \begin{minipage}{0.72\linewidth}
      \centering
      \includegraphics[width=\linewidth]{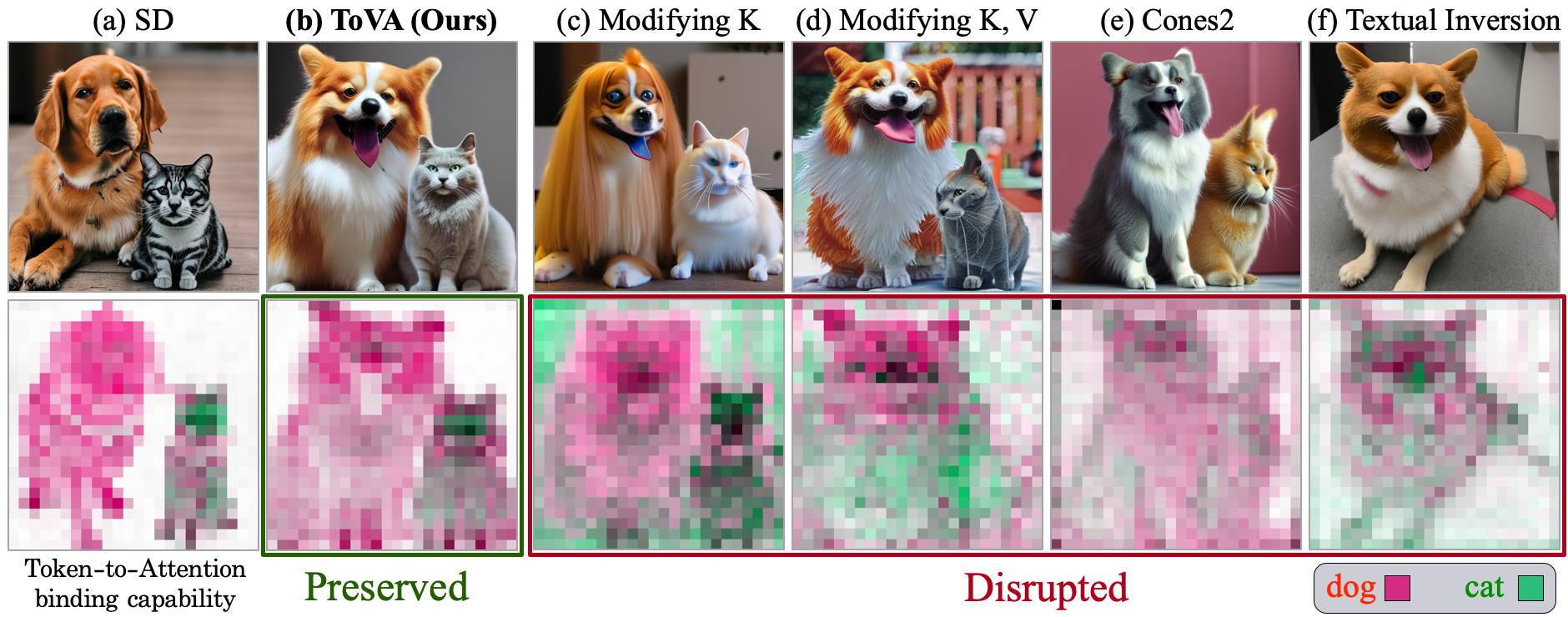}
        \caption{\textbf{Visualization comparison of attention maps.} Preserving the value (b) maintains attention clearly focused on objects and background, whereas modifying the key (c, d, e, f) disrupts attention distributions, compared to Stable Diffusion (a).}
      \label{fig:disruption}
    \end{minipage}
  \end{tabular}
\end{figure*}

Let us consider a set of concepts $C = \{C_1, C_2, \ldots, C_{I}\}$, where each concept $C_i$ is associated with a dedicated adapter $\mathcal{A}_{C_i}$ that is trained to learn its respective projection. Then we can construct a merged adapter for multi-concept personalization by the following merging operation:
\begin{equation}
\mathcal{A}_m(\textbf{x}) = \sum_{i=1}^{I} \boldsymbol{\lambda}_i \odot \mathcal{A}_{C_i}(\textbf{x})
\end{equation}
where $\mathcal{A}_m$ denotes the merged adapter, $\odot$ represents the Hadamard product, and $\boldsymbol{\lambda}_i$ is a weight tensor that scales the element-wise contribution of each adapter’s output, allowing them to be amplified or suppressed.
For the case in LoRA, this can be achieved by:
\begin{equation}
(\mathbf{B}_m\mathbf{A}_m)(\textbf{x}) = \sum_{i=1}^{I} \boldsymbol{\lambda}_i \odot (\mathbf{B}_i\mathbf{A}_i)(\textbf{x})
\end{equation}
where $\mathbf{A}_i$ and $\mathbf{B}_i$ are the weight matrices of the LoRA adapter for concept $C_i$.

Consider using this merged adapter to value projection $W^v$ of cross-attention in U-Net. Given a set of tokens $S = \{s_1, s_2, \dots, s_{n}\}$, a text feature ${\mathbf{c} \in \mathbb{R}^{n \times d}}$ computed by $\mathbf{c} = E(S)$ represents a matrix composed of text embedding vectors corresponding to each token. Here, $n$ is the number of tokens, $d$ is the dimensionality of the embedding space, and $E(\cdot)$ is the text encoder.
Then the value is updated using the merged adapter as:
\begin{equation}
\mathbf{V} = W^v (\mathbf{c}),
\quad
\mathbf{V}_m = \mathcal{A}_m(\mathbf{c}),
\quad
\end{equation}
\begin{equation}
\mathbf{V'} = \mathbf{V} + \mathbf{V}_m.
\end{equation}
From the above equations, this merging process inevitably mixes the specific features learned via different adapters, resulting in the concept mixing problem (Figure \ref{fig:teaser} (a)). Therefore, a new merging-free adaptation method is required to effectively prevent concept mixing.

To this end, we design a novel method that supports multi-concept inference without merging adapters at inference time. Suppose a user wants to personalize multi concepts to the corresponding tokens among a set of tokens $S$. To modify the value $\mathbf{V}_i$ for a specific token $s_i$ with concept $C_i$, we feed only that token's embedding $\mathbf{c}_i = E(s_i)$ into the corresponding adapter as $\mathcal{A}_{C_i}(\mathbf{c}_i)$. We then compute the updated value $\mathbf{V}'$ as follows:
\begin{align}
    \begin{split}
        \mathbf{V} &= W^v (\mathbf{c}),
        \quad\mathbf{V}_i = \mathcal{A}_{C_i}\bigl(\mathbf{c}_i\bigr),\\
        \boldsymbol{\delta}_i &=
        \begin{bmatrix}
        0 & \cdots & 1 & \cdots & 0
        \end{bmatrix}
        \in \mathbb{R}^{1 \times n},\\
        \mathbf{V}' &= \mathbf{V} + \boldsymbol{\delta}_i^\top\mathbf{V}_i,
    \end{split}
\end{align}
where $\boldsymbol{\delta}_i$ is a one-hot vector selecting which row of $\mathbf{V}\in \mathbb{R}^{n\times v}$ should be added with $\mathbf{V}_i\in \mathbb{R}^{1\times v}$, and $v$ is the output dimension. As a result, concept-specific adapters for the corresponding tokens are applied independently, eliminating the need for adapter merging and avoiding unwanted mixing of concepts (see Figure \ref{fig:mainfigure} (a)). Furthermore, at the inference, the multiple adapters can modify corresponding values token-wisely (Figure \ref{fig:mainfigure} (b)) as follows:
\begin{equation}
\mathbf{V}' = \mathbf{V} + \sum_{i=1}^K \boldsymbol{\delta}_i^\top \mathcal{A}_{C_i}\bigl(\mathbf{c}_i\bigr),
\end{equation}
where $K$ is the number of tokens to be conceptualized. In this paper, we employ LoRA \cite{hu2021lora} as the adapter for all of our experiments. Note that our approach is independent of the adapter structure, which means other adapters can be used for ToVA.

\begin{figure}[t]
    \centering
    \includegraphics[width=1\linewidth]{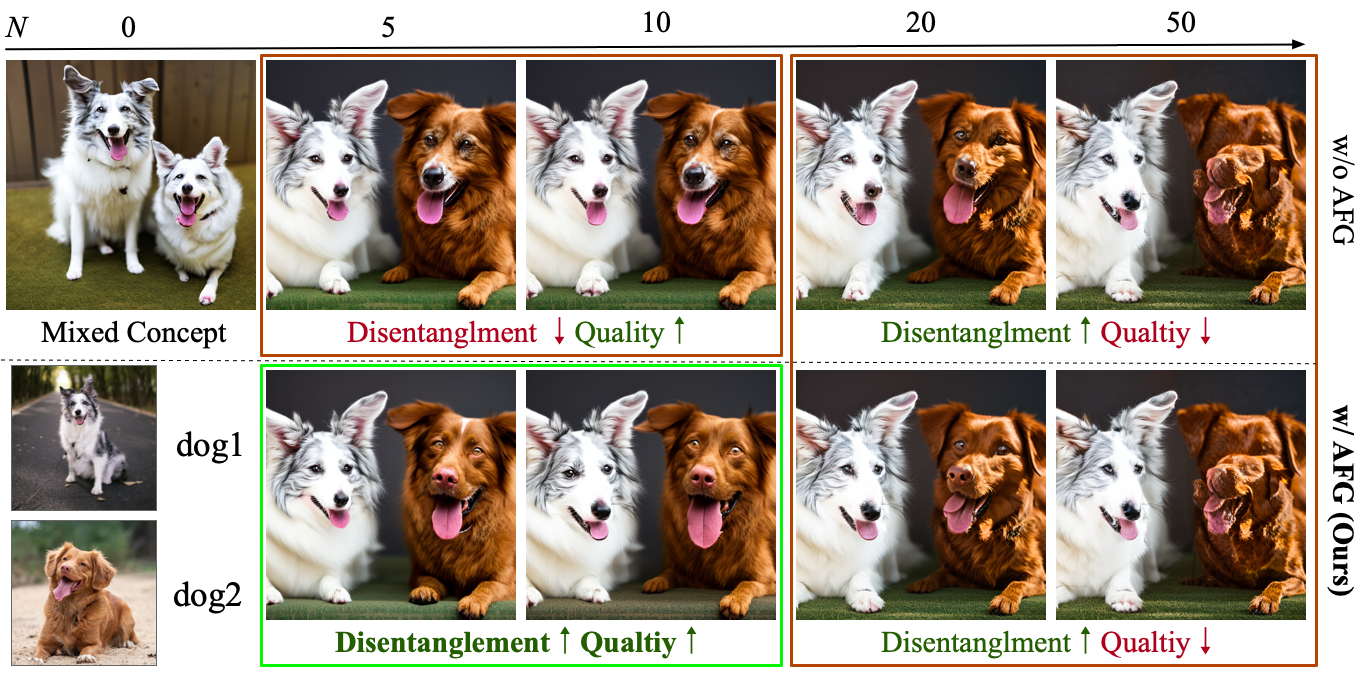}
    \caption{\textbf{Tradeoff between disentanglement and image quality.} The parameter $N$ represents the Stage 1 threshold at which latent optimization is halted. This threshold highlights a tradeoff between concept disentanglement and image quality. To mitigate this issue, we introduce Attention-Fixing Guidance (AFG).}
    \label{fig:AFGabl}
\end{figure}

\subsection{Why Is Only Value Projection Trained?}
\label{sec:attentiondisruption}
While the cross-attention layer in U-Net employs both key and value components derived from text embeddings, we only modify the value component to preserve the token-attention binding capacity of Stable Diffusion. We empirically found that modifying keys together degrades this capacity, as evidenced by the disrupted and noisy attention patterns in Figure \ref{fig:disruption} (c). Typically, the T2I models like Stable Diffusion refine their attention progressively during denoising: initially, the attention is scattered (high entropy), and it becomes increasingly focused (low entropy) over time, effectively binding each token to a specific region in the latent space (see Figure \ref{fig:LODAabl}).
We further note that existing merging-free training methods such as Cones2 \cite{liu2023cones2} and Textual Inversion \cite{gal2022textualinversion} can weaken this token-attention binding mechanism, leading to concept mixing and degrading performance in multi-concept scenarios. We hypothesize that this occurs because these methods directly optimize text embeddings, which then feed into the key projection for cross-attention; since the attention map (or attention weight) is computed via $\mathbf{Q}\mathbf{K}^T$, changes to $\mathbf{K}$ risk disrupting the attention map—particularly in few-shot personalization.

To verify this, we measure how attention-map entropy changes across denoising steps for each method, including variations that modify only the key, only the value, or the both. Figure \ref{fig:disruption} confirms that modifying the key—either alone or alongside the value—introduces noisy, disrupted attention maps. Notably, as shown in Figure \ref{fig:entropy}, such key-modifying methods keep overall attention entropy at roughly the same level rather than reducing it. Based on our empirical analysis, we propose ToVA that selectively trains only the value projection while freezing the key component, preserving the token-attention binding capacity of the pre-trained model.

\subsection{Inference with Latent Optimization}
\label{sec:LODA}
\label{loda}
As illustrated in Figure \ref{fig:teaser} (b), concept mixing arises from the entangled attention—one of the core issues we aim to resolve. While Attend-and-Excite \cite{chefer2023attendandexcite} pioneered latent optimization to boost attention intensity on specific tokens, our approach takes a distinct direction. Instead of merely enhancing intensity, we diversify the distribution of each attention token. We achieve this through a two-phase process: (1) we optimize the latent space to separate the attention distributions, and (2) we fix this attention to maintain their disentanglement. As identified by Ho et al.\cite{ho2020denoising}, denoising process can be distinct into semantic (structural component) and perceptual (fine-grained refinement) stage. Significantly, we restrict latent optimization to the semantic stage, as applying it during the perceptual stage could compromise image quality (see Figure \ref{fig:AFGabl}).
\subsubsection{Stage 1: Latent Optimization}
Let $\mathcal{P}$ be the input prompt, and let $S = \{s_1, s_2, \dots, s_k\}$ be a subset of token indices that a user wants to modify (i.e., each component refers to an index of token). During denoising, we extract the cross-attention maps at a $24 \times 24$ resolution from each attention layer and head, then average them. The resulting aggregated map $A_t \in \mathbb{R}^{24\times 24\times k}$ contains $k$ attention maps, one for each token in $S$. These maps are then smoothed and normalized to form probability distributions, yielding $P_t^1, P_t^2, \dots, P_t^k$, where $P\in \mathbb{R}^{24\times24}$ with a timestep $t$. Then we compute the Kullback-Leibler (KL) divergence for each pair of tokens in $S$ as follows:
\begin{equation}
\text{KL}_t^{(i,j)}
\;=\;
\sum_{m=1}^{24}\sum_{n=1}^{24} P_t^i[m,n] \,\log~\!\Bigl(\frac{P_t^i[m,n]}{P_t^j[m,n]}\Bigr)
\end{equation}
which quantifies the relationship between the distributions of two tokens. Then we compute the harmonic mean ($\mathrm{HM}$) of each $\text{KL}_t^{(i,j)}$ as follows: 
\begin{equation}
\text{KL}^H_t
\;=\;
\mathrm{HM}\bigl(\{\text{KL}_t^{(i,j)} \mid i,j \in S,\; i \neq j\}\bigr).
\end{equation}
Finally, as our goal is to disentangle the attention, the distribution of each token's attention has to be diverge. Therefore, the form of loss can be achieved by passing this $\text{KL}^H_t$ to negative ReLU, then the latent can be update as:
\begin{align}
    \begin{split}
    \mathcal{L}_{KL} = ReLU(\tau-\text{KL}^H_t), \\
    \mathbf{z}_t' \leftarrow \mathbf{z}_t - \eta_t \nabla_{\mathbf{z}_t} \mathcal{L}_{KL},
    \end{split}
\end{align}
where $\tau$ and $\eta_t$ denote the ReLU threshold and the gradient update step size, respectively. Here, $\tau$ acts as a safeguard: if the divergence exceeds $\tau$, the gradient becomes zero to prevent instability. After updating latent $z_{t}'$, denoising is performed with the updated latent which implies the diverged attention of each token. We run this latent-optimization stage for up to $N$ steps, where we use $N=10$ for all experiments (see Figure \ref{fig:AFGabl}).
\begin{figure*}[t]
    \centering
    \includegraphics[width=1\linewidth]{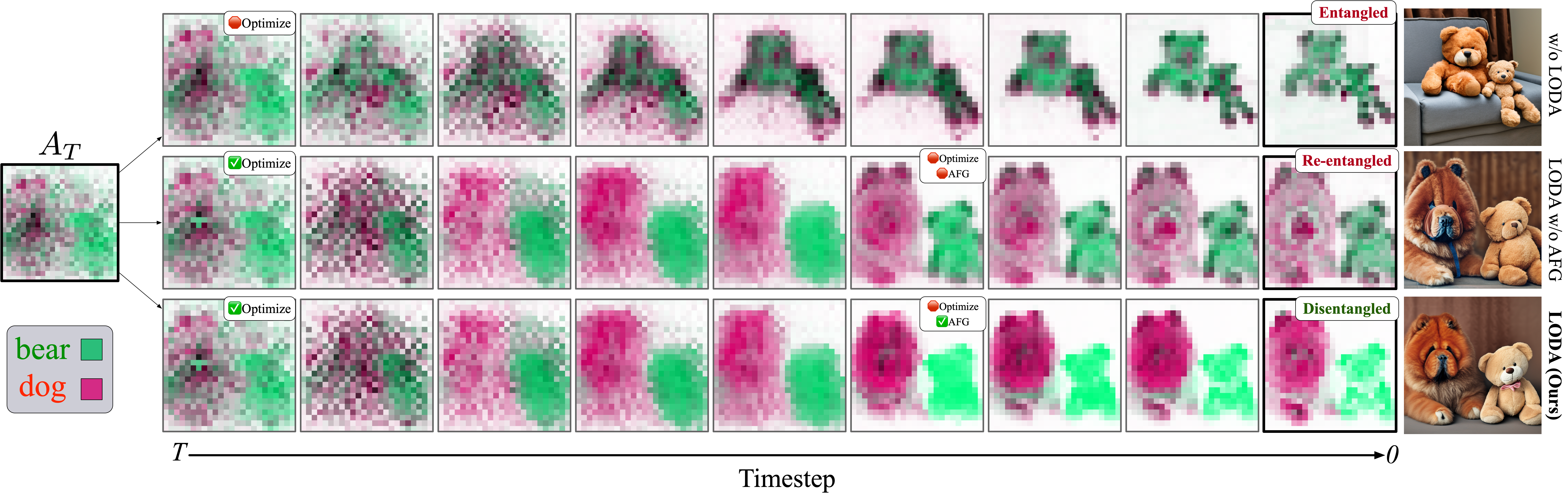}
\caption{\textbf{Visual analysis of attention maps demonstrating the effect of LODA.} \underline{\textit{First row}}: Stable Diffusion occasionally makes entangled attention, leading to mixed concept. \underline{\textit{Second row}}: LODA addresses this issue by performing latent optimization to disentangle the attention maps; nevertheless, if the optimization is early halted, the attention maps quickly revert to an entangled state. \underline{\textit{Third row}}: to resolve this, we introduce Attention Fixing Guidance, employing a split-and-preserve strategy that maintains the disentangled attention structure throughout the denoising process. All results are visualized from same latent and seeds.}
    \label{fig:LODAabl}
\end{figure*}

\subsubsection{Stage 2: Attention Fixing Guidance}
As shown in Figure \ref{fig:AFGabl}, we observed that continuing optimization at the perceptual stage can degrade image quality. On the other hand, stopping the latent optimization too early results in concept mixing, suggesting a trade-off between concept disentanglement and image quality when relying solely on latent optimization. To address this issue, we introduce a new stage, Attention Fixing Guidance (AFG), to preserve disentangled attention. As shown in Figure \ref{fig:LODAabl}, once latent optimization ceases, the attention quickly begins to “re-entangle”, causing mixed concepts. AFG counters this effect by “fixing” attention via an additional guidance step. Concretely, we first compute a Gaussian-smoothed attention map for each token of index $i\in S$ as follows:
\begin{align}
    A_t^i = \mathrm{Gaussian}\bigl(A_t[:,:,i]\bigr),
\end{align}
where $A_t\in \mathbb{R}^{24\times24\times k}$. Then for each token index $i$ in $S$, we compute threshold $\theta_i$ with a percentile hyperparameter $\gamma$,
\begin{align}
\theta_i \;=\; \mathrm{Percentile}\bigl(A_t^i,\; \gamma \bigr).
\end{align}
This Percentile operation means that if $\gamma = 0.9$, $\theta_i$ represents the threshold value corresponding to the 90-th percentile of element in $A^i_t$. Using this threshold, we can calculate an attention mask as follows:
\begin{equation}
M_i[m,n]=\begin{cases} 1 & \text{if } A_t^i[m,n] \geq \theta_i\\ 0 & \text{otherwise} \end{cases}.
\end{equation}
This mask is attached to the attention forwarding process of U-Net in Stable Diffusion as follows:
\begin{align}
    \begin{split}
        z_{t-1} &\leftarrow {SD(z_t,\mathcal{P},t,p,m)},\\
        {A_t^i}'&
        \;\leftarrow
        A_t^i  + p \cdot M^i +
        m \cdot\sum^S_{\substack{j \neq i}} 
         M^j,  \ \forall (i,j) \in S,
    \end{split}
\end{align}
where $p$ amplifies the attention on the target token $A^i$ by adding $M^i$, while $m$ ensures other token's highlighted region $M^j$ to not affect the target token. This formula ensures that each token's highlighted attention region remains for itself while reducing the influence of other tokens. Through this Attention-Fixing Guidance, we ensure that the disentangled attention remains focused solely on the designated position, effectively maintaining both attention disentanglement and image quality. For our experiments, we set $p=3$ and $m=-1e8$. The complete algorithm for our latent optimization approach is provided in Supplementary D.



\begin{figure*}[t]
    \centering
    \includegraphics[width=1\linewidth]{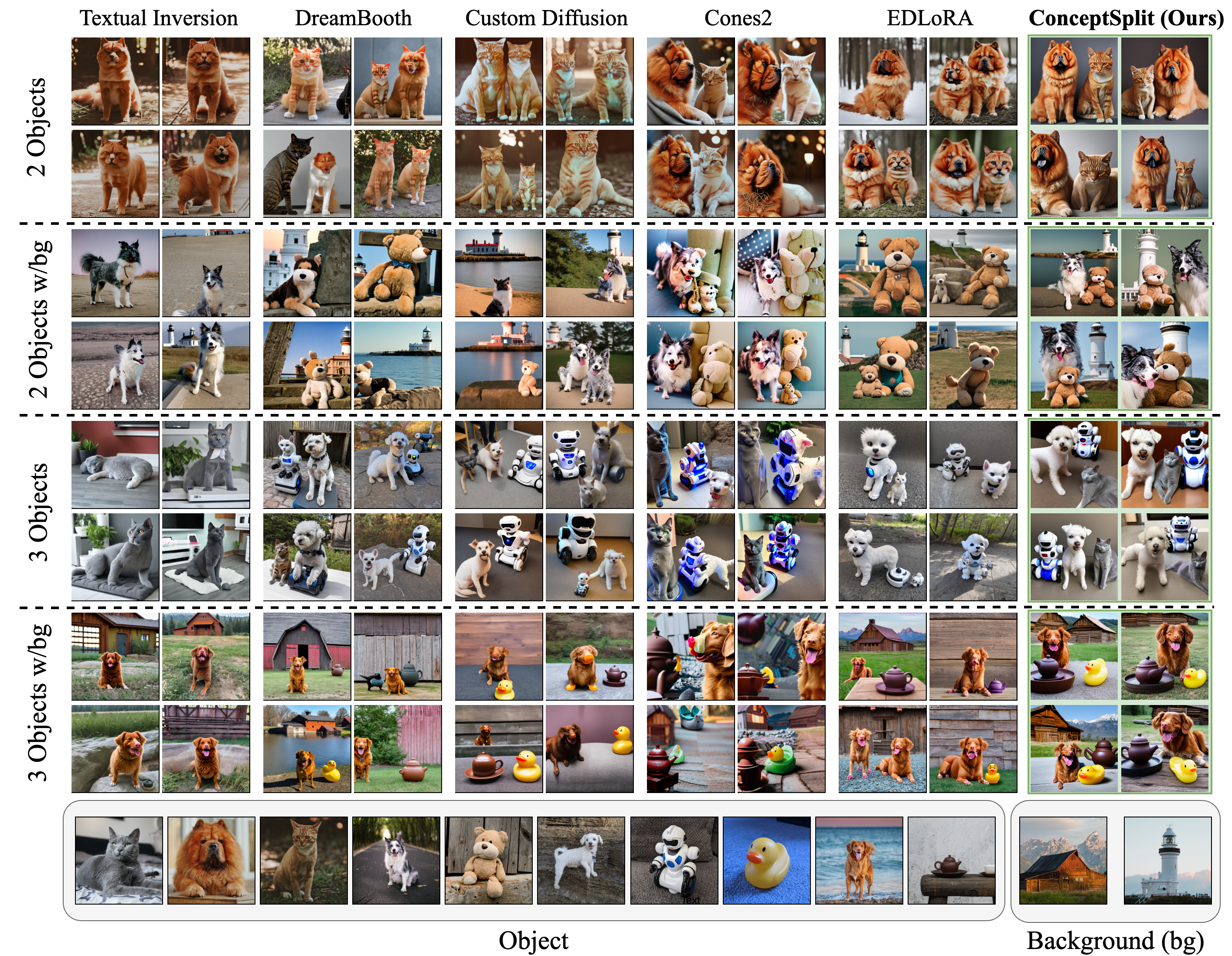}
    \caption{\textbf{Qualitative comparison on Stable Diffusion 2.1.} In multi-object scenarios, our approach demonstrates improved separation between individual objects and more distinct disentanglement of learned concepts, while other methods suffer from mixed concepts. Qualitative comparison for single-object scenarios is in Supplementary E.}
    \label{fig:quality}
\end{figure*}

\section{Experiments}
\textbf{Dataset and Experimental Setup.} We utilize the dataset from previous studies~\cite{ruiz2023dreambooth, choi2023custom, gal2022textualinversion}, specifically focusing on personalization of both objects and backgrounds. While background highly affect metrics in TA and IA, we organize our experiments into two categories: objects with backgrounds and without. For each category, we systematically selected representative combinations of objects and backgrounds. We then designed 10 distinct prompts, such as ``A photo of a \{object\} sitting next to a \{object\}, with a background of \{background\}." Each prompt generated 50 images, leading to 500 images per combination.

\noindent\textbf{Evaluation Metrics.} We evaluate our method using four key metrics: Text-Alignment (TA)~\cite{radford2021clip}, CLIP-based Image-Alignment (C-IA)~\cite{radford2021clip}, DINO-based Image-Alignment (D-IA)~\cite{caron2021emerging}, and GenEval (GE)~\cite{ghosh2024GenEval}. TA measures the CLIP similarity between the generated image and the input prompt, while C-IA and D-IA measure the similarity of the generated subjects against reference images using CLIP and DINO, respectively.
TA and IA scores may not be sufficient for judging compositional success because they do not verify the presence and separability of all concepts described in a prompt. An image might score highly on such metrics even if it suffers from mixed concepts. To overcome this, we incorporate GenEval (GE), a metric designed specifically to assess compositional correctness. GE leverages an object-detection model to verify whether each object from the prompt appears in the generated image. It assigns a binary score of 1 if all objects are present, 0 otherwise. We report the average score across all generated images.

\noindent\textbf{Baselines.} We compare our method with five baseline methods: DreamBooth~\cite{ruiz2023dreambooth}, Textual Inversion~\cite{gal2022textualinversion}, Custom Diffusion~\cite{choi2023custom}, Cones2~\cite{liu2023cones2}, a textual-embedding-based method, and EDLoRA~\cite{gu2024mixofshow}, an adapter-based technique proposed in Mix-of-Show. Further implementation details are provided in Supplementary A.

\subsection{Main Results}
\noindent\textbf{Quantitative Results.}
As shown in Table~\ref{table:maintable}, a clear limitation of existing methods is their poor performance on the GE metric, which points to the frequent occurrence of concept mixing in their results. While they may achieve high scores in metrics like C-IA, this often comes at the cost of compositional correctness. In contrast, our approach not only maintains competitive TA and C-IA scores but also significantly outperforms all baselines in both D-IA and GE. This dual superiority demonstrates its effectiveness in simultaneously separating concepts and preserving their fine-grained visual features.

\noindent\textbf{Qualitative Results.}
Figure \ref{fig:quality} presents qualitative comparisons in multi-object scenarios. Baselines like Textual Inversion~\cite{gal2022textualinversion}, DreamBooth~\cite{ruiz2023dreambooth}, and Custom Diffusion~\cite{choi2023custom} commonly suffer from object omission or concept mixing with semantically related objects. Cones2~\cite{liu2023cones2}, which utilizes layout guidance on a region of the attention map during inference, fails when a background is added (e.g., the bear is degraded in the second row, and the dog and robot are mixed in the third). EDLoRA~\cite{gu2024mixofshow}, which fuses trained LoRA weights to incorporate knowledge of multiple concepts, also exhibits concept omission or mixing (e.g., no cat in the second row, mixed dog and robot in the third). In contrast, our method consistently generates images with distinct, high-fidelity representations for all concepts.

\begin{table}[t]
\centering
\setlength{\tabcolsep}{5pt} 
\renewcommand{\arraystretch}{1} 
\resizebox{0.48\textwidth}{!}{ 
\begin{tabular}{clcccccccc}
\toprule
\multirow{2}{*}{\textbf{\# Objs.}} & \multirow{2}{*}{\textbf{Methods}} & \multicolumn{4}{c}{\textbf{w/o Background}} & \multicolumn{4}{c}{\textbf{w/ Background}}
\\ \cmidrule(lr){3-6}\cmidrule(lr){7-10}
& & \textbf{TA} & \textbf{C-IA}& \textbf{D-IA} & \textbf{GE}& \textbf{TA} & \textbf{C-IA}& \textbf{D-IA} & \textbf{GE}\\
\midrule
\multirow{6}{*}{\shortstack[l]{1 Obj.}} 
& Textual Inversion \cite{gal2022textualinversion}&0.231&0.843&0.677&\underline{0.946}&0.223&0.707&0.416&0.968\\
& DreamBooth \cite{ruiz2023dreambooth} & \underline{0.234} & 0.838 & 0.723 & 0.916 & 0.224 & 0.684 & 0.415 & 0.952 \\
& Custom Diffusion \cite{kumari2023custom} & 0.222 &\textbf{0.873} & \underline{0.800} & \textbf{0.954} & 0.221 & \underline{0.712} & 0.450 & 0.950 \\
& Cones2 \cite{liu2023cones2} & \textbf{0.235} & 0.812 & 0.609 & 0.926 & \underline{0.249} & 0.692 & 0.414 & 0.864 \\
& EDLoRA \cite{gu2024mixofshow}& 0.221 & \underline{0.866} & 0.778 & 0.938 & 0.241 & \textbf{0.722} & \underline{0.509} & \textbf{0.998} \\
& \textbf{ConceptSplit (Ours)} & 0.229 & 0.838 & \textbf{0.809} & 0.942 & \textbf{0.286} & 0.705 & \textbf{0.590} & \underline{0.972} \\

\midrule
\multirow{6}{*}{\shortstack[l]{2 Objs.}} 
& Textual Inversion \cite{gal2022textualinversion}&0.236&0.722&0.482&0.536&0.244&0.677&0.377&0.002\\
& DreamBooth \cite{ruiz2023dreambooth} & \textbf{0.244} & 0.722 & 0.447 & \underline{0.624} & \underline{0.251} & 0.660 & 0.307 & 0.204 \\
& Custom Diffusion \cite{kumari2023custom} & 0.220 & 0.748 & 0.402 & 0.005 & \underline{0.251} & 0.650 & 0.366 & 0.135 \\
& Cones2 \cite{liu2023cones2} & 0.217 & \underline{0.772} & 0.514 & 0.582 & 0.217 & 0.686 & 0.334 & \underline{0.288} \\
& EDLoRA \cite{gu2024mixofshow}& 0.201 & \textbf{0.776} & \underline{0.566} & 0.342 & 0.218 & \textbf{0.705} & \underline{0.420} & 0.237 \\
& \textbf{ConceptSplit (Ours)} & \underline{0.238} & 0.761 & \textbf{0.759} & \textbf{0.902} & \textbf{0.282} & \underline{0.687} & \textbf{0.573} & \textbf{0.648} \\

\midrule
\multirow{6}{*}{\shortstack[l]{3 Objs.}} 
& Textual Inversion \cite{gal2022textualinversion}& \underline{0.261}&0.672&0.432&0.016&\underline{0.279}&0.650&0.355&0.009\\
& DreamBooth \cite{ruiz2023dreambooth} & 0.225 & \underline{0.720} & 0.405 & 0.001 & 0.238 & \underline{0.674} & 0.298 & 0.001 \\
& Custom Diffusion \cite{kumari2023custom} & 0.235 & 0.684 & 0.361 & 0.027 & 0.244 & 0.609 & 0.326 & \underline{0.120} \\
& Cones2 \cite{liu2023cones2} & 0.230 & \textbf{0.759} & 0.440 & \underline{0.284} & 0.231 & \textbf{0.690} & 0.317 & 0.080 \\
& EDLoRA \cite{gu2024mixofshow}& 0.243 & 0.710 & \underline{0.497} & 0.095 & 0.263 & 0.669 & \underline{0.393} & 0.004 \\
& \textbf{ConceptSplit (Ours)} &\textbf{0.271} & 0.717 & \textbf{0.669} & \textbf{0.344} & \textbf{0.301} & 0.673 & \textbf{0.557} & \textbf{0.384} \\
\bottomrule
\end{tabular}
}
\caption{\textbf{Quantitative comparison.} Our method excels at multi-object generation, outperforming baselines across key metrics like Text-Alignment (TA), DINO-based Image-Alignment (D-IA), and GenEval (GE). It also remains competitive in single-object tasks, showcasing its robustness and scalability.}
\label{table:maintable}
\end{table}

\subsection{Ablation Studies}
\noindent\textbf{Ablation Study on ToVA.}
Table \ref{table:tova} presents our ablation study for ToVA. As proposed in Section \ref{sec:tova}, ToVA modifies only the value projection. We compare this approach against variants modifying the key alone, or both the key and value, thereby validating our design choice. Our experiments demonstrate that modifying only the value yields superior IA scores in both CLIP and DINO, whereas the other approaches result in degraded images (Supplementary C).

\noindent\textbf{Ablation Study on LODA.}
Table~\ref{table:loda} presents our ablation study on LODA’s two-stage inference (Section~\ref{sec:LODA}). Evaluating each stage, Latent Optimization (Stage 1) improves two-object generation but struggles beyond two Objs. Attention Fixing Guidance (Stage 2) consistently enhances results across all scenarios. We also compared LODA to Attend-and-Excite (AaE) on vanilla Stable Diffusion; LODA generates distinct object representations, while AaE struggles with feature blending (Supplementary G).
\begin{table}[t]
\centering
\setlength{\tabcolsep}{4pt} 
\renewcommand{\arraystretch}{1} 
\resizebox{0.48\textwidth}{!}{
\begin{tabular}{ccccccccccc} 
\toprule
\multirow{2}{*}{\textbf{\# Objs.}} & \multicolumn{2}{c}{\textbf{ToVA}} & \multicolumn{4}{c}{\textbf{w/o Background}} & \multicolumn{4}{c}{\textbf{w/ Background}}
\\ \cmidrule(lr){2-3}\cmidrule(lr){4-7}\cmidrule(lr){8-11} 
& \textbf{Key} & \textbf{Value} & \textbf{TA} & \textbf{C-IA} & \textbf{D-IA} & \textbf{GE} & \textbf{TA} & \textbf{C-IA} & \textbf{D-IA} & \textbf{GE} \\ 

\midrule
\multirow{3}{*}{\shortstack[l]{1 Obj.}} 
&  \checkmark &  \checkmark  &\textbf{0.237}&0.716	& 0.446&0.940	&\textbf{0.292}	&0.657 & 0.336&0.970 \\
& \checkmark & -  &0.236	&0.715	& 0.175&0.910	&0.268	&0.669 & 0.197&\textbf{0.980}\\
& - & \checkmark &0.229	&\textbf{0.838} & \textbf{0.809}&\textbf{0.942}	&0.286	&\textbf{0.705} & \textbf{0.590}&0.972 \\

\midrule
\multirow{3}{*}{\shortstack[l]{2 Objs.}} 
&  \checkmark &  \checkmark  &\textbf{0.247}&	0.673& 0.474&	0.480&	\textbf{0.289}&	0.628& 0.331&	\textbf{0.520 }\\
& \checkmark & -  &0.235	&0.652& 0.300&	\textbf{0.560}&	0.252&	0.617& 0.225&	0.390\\
& - & \checkmark & 0.220& \textbf{0.756}& \textbf{0.566}&	0.550	&0.274&	\textbf{0.688}& \textbf{0.360}&	0.052 \\

\midrule
\multirow{3}{*}{\shortstack[l]{3 Objs.}} 
&  \checkmark &  \checkmark  &0.257&	0.653& 0.433&	0.002&	0.280&	0.622	& 0.333&0.004 \\
& \checkmark & -  &0.230	&0.644& 0.343&	0.002	&0.247	&0.611	& 0.270&0.010\\
& - & \checkmark & \textbf{0.267} &\textbf{0.702}& \textbf{0.455}&	\textbf{0.012}	&\textbf{0.297 }&\textbf{0.663}	& \textbf{0.360}&\textbf{0.074} \\

\bottomrule
\end{tabular}}

\caption{\textbf{Ablation study on ToVA.} Note that to solely evaluate the effectiveness of ToVA, LODA was intentionally excluded from this ablation. Only value training shows most well image-aligned scores in both CLIP and DINO, while other shows lower alignment scores result in degraded images.}
\label{table:tova}
\end{table}

\begin{table}[t]
\centering
\setlength{\tabcolsep}{4pt}
\renewcommand{\arraystretch}{1}
\resizebox{0.48\textwidth}{!}{
\begin{tabular}{ccccccccccc} 
\toprule
\multirow{2}{*}{\textbf{\# Objs.}} & \multicolumn{2}{c}{\textbf{LODA}} & \multicolumn{4}{c}{\textbf{w/o Background}} & \multicolumn{4}{c}{\textbf{w/ Background}}
\\ \cmidrule(lr){2-3}\cmidrule(lr){4-7}\cmidrule(lr){8-11} 
& \textbf{Stage 1} & \textbf{Stage 2} & \textbf{TA} & \textbf{C-IA} & \textbf{D-IA} & \textbf{GE} & \textbf{TA} & \textbf{C-IA} & \textbf{D-IA} & \textbf{GE} \\ 

\midrule
\multirow{3}{*}{\shortstack[l]{2 Objs.}} 
& - & -  & 0.220& 0.756 & 0.566 & 0.550 &0.274 & 0.688 & 0.360 & 0.052 \\
& \checkmark & -  & 0.235 & 0.566 & 0.574 & 0.825 & 0.277 &\textbf{0.688} & 0.368 & 0.162\\
& \checkmark & \checkmark & \textbf{0.238} & \textbf{0.761} & \textbf{0.809} & \textbf{0.902} & \textbf{0.282} & 0.687 & \textbf{0.590} & \textbf{0.648} \\

\midrule
\multirow{3}{*}{\shortstack[l]{3 Objs.}} 
& - & -  & 0.267 & 0.702 & 0.455 & 0.012 & 0.297 &0.663 & 0.360 & 0.074 \\
& \checkmark & -  &0.264& 0.704 & 0.469 & 0.026 & 0.300 & 0.671 & 0.373 & 0.074\\
& \checkmark& \checkmark &\textbf{0.271} & \textbf{0.702} & \textbf{0.669} & \textbf{0.344} & \textbf{0.301}& \textbf{0.673} & \textbf{0.557} & \textbf{0.384}\\

\bottomrule
\end{tabular}}
 \caption{\textbf{Ablation study on LODA.} Applying Stage 1 improves effectiveness in 2-object scenarios but yields no improvement for 3-object cases. With Stage 2, overall performance increases substantially across all metrics, particularly on D-IA and GE, demonstrating robust visual alignment and concept disentanglement.}
\label{table:loda}
\end{table}

\section{Conclusion}
We introduce ConceptSplit, a framework for decoupled multi-concept personalization in T2I diffusion models. To mitigate concept entanglement, we propose Token-wise Value Adaptation (ToVA) for selective value projection and Latent Optimization for Disentangled Attention (LODA) to enhance concept separation. ConceptSplit effectively reduces interference in multi-object scenarios. However, its performance is limited by the pre-trained diffusion model, constraining generalization. Future work should address this dependency to improve multi-concept representation.
\maketitlesupplementary

\renewcommand{\thesection}{\Alph{section}}

\setcounter{section}{0}
\setcounter{figure}{0}
\setcounter{table}{0}
\setcounter{algorithm}{0}

\begin{algorithm}[t]
    \caption{Denoising steps with Latent Optimization for Disentangled Attention (LODA)}
    \label{alg:loda}
    \begin{algorithmic}[1]
        \State \textbf{Input:} initial latent $z_T$,
        prompt $\mathcal{P}$,
        set of token indices $S$,
        set of timesteps $t=\{T,\dots,0\}$,
        threshold $\gamma$,
        Stage1 end step $N$,
        Stable Diffusion model $\operatorname{SD}$.
        \State \textbf{Output:} Denoised latent $z_0$
        \State $\mathrm{step} \gets 0$
        \For{$t = T$ down to $0$}
            \If{$\mathrm{step} < N$}
                \State $A_t \leftarrow \operatorname{SD}\bigl(z_t, \mathcal{P}, t\bigr)$
                
                \For{each token index $i$ in $S$}
                    \State $A_t^i \leftarrow A_t[:,:,i]$
                    \State $A_t^i \leftarrow Gaussian\bigl(A_t^i\bigr)$
                    \State $P_t^i \leftarrow Normalize\bigl(A_t^s\bigr)$
                \EndFor
                \State $
                        \text{KL}^H_t
                        \leftarrow
                        \mathrm{HM}\bigl(\{\text{KL}_t^{(i,j)} \mid i,j \in S,\; i \neq j\}\bigr)
                        $
                \State$\mathcal{L}_{KL} \leftarrow ReLU(\gamma-\text{KL}^H_t)$
                \State $z_t' \gets z_t - \eta_t \,\nabla_{z_t} \,\mathcal{L}_{KL}$
            \EndIf
            \State $z_{t-1} \leftarrow \operatorname{SD}\bigl(z_t', \mathcal{P}, t\bigr)$
            \State ${\mathrm{step} \leftarrow \mathrm{step} + 1}$
        \EndFor
        
        \State \Return $z_0$
    \end{algorithmic}
\end{algorithm}

\begin{table}[h]
\centering
\begin{tabular}{ccc}
\hline
\textbf{Method} & \textbf{Dog} & \textbf{Cat} \\ 
\hline
SD & -0.003600 & -0.004531 \\
ToVA &-0.003041	&-0.005061\\
Modifying K & -0.000266 & -0.000280 \\
Modifying K,V & -0.000359 & -0.000176 \\
Cones2 & -0.000385 & -0.000421 \\
Textual Inversion & -0.000933 & -0.001004 \\
\hline
\end{tabular}
\caption{
\textbf{Average entropy change of attention maps (\(\Delta \mathcal{H}\)) across diffusion steps.}
We extract attention maps corresponding to the tokens ``\textit{cat}" and ``\textit{dog}" from the prompt ``A photo of a dog sitting next to a cat".
}
\label{tab:entropy_change}
\end{table}

\section{Experimental Detail}
We used the RTX3090 graphic card for training and inference. For inference, we used DDIM scheduler \cite{song2020ddim} with 50 steps and 7.5 classifier-free guidance weight \cite{ho2021classifier}. We use Stable Diffusion v2-1\footnote{\url{https://huggingface.co/stabilityai/stable-diffusion-2-1}} with 768x768 resolution as the pre-trained model.

\label{sec:Experimentaldetails}
\noindent\textbf{Textual Inversion~\cite{gal2022textualinversion}.}
We use the third-party implementation of huggingface~\cite{von-platen-etal-2022-diffusers} for Textual Inversion. We train each setting with a learning rate of $5 \times 10^{-4}$, step size of 3000, and batch size of 4.

\noindent\textbf{DreamBooth~\cite{ruiz2023dreambooth}.}
We use the third-party implementation of huggingface~\cite{von-platen-etal-2022-diffusers} for Dreambooth. We train each setting with a learning rate of $1 \times 10^{-5}$, step size of $400 \times$ number of subjects, and batch size of 1. Prior preservation loss was used, and a loss weight of 1 was used. 200 class images were used for prior preservation loss.

\noindent\textbf{Custom Diffusion~\cite{kumari2023custom}.}
We use the official implementation of custom diffusion\footnote{\url{https://github.com/adobe-research/custom-diffusion}}. We train each setting with a learning rate of $1 \times 10^{-4}$, step size of $500 \times$ number of subjects, and batch size of 1. Generated images shown in our paper were trained with prior preservation loss that prevents the leak of personalized concepts when generating other concepts in diffusion models. Loss weight of 1 was used for prior preservation loss. 200 class images were used for prior preservation loss.

\noindent\textbf{Cones2~\cite{liu2023cones2}.}
We use the official implementation for Cones2\footnote{\url{https://github.com/ali-vilab/Cones-V2}}. We train each setting with a learning rate of $1 \times 10^{-4}$, step size of 1500, and batch size of 1. With prompt regularization. We implement layout guidance for inference 2 objects and 3 objects. 

\noindent\textbf{EDLoRA~\cite{gu2024mixofshow}.}
We use the official implementation for EDLoRA\footnote{\url{https://github.com/TencentARC/Mix-of-Show.git}}. We implement this code to SD2.1 for comparison. We train each EDLoRA with  setting with learning rate of $1\times10^{-3}$ for text embedding, $1\times10^{-5}$ for text encoder and $1\times10^{-4}$ for unet. We set every rank of LoRA to 4. We set the alpha value for gradient fusion to 1 for the U-net and text encoder.

\noindent\textbf{ConceptSplit (Ours)}

\noindent\textbf{ToVA.} We used LoRA with rank 64 for our ToVA, and we used prompt regularization, as proposed in Cones2~\cite{liu2023cones2} We utilized 200 prompts using ChatGPT~\cite{radford2018improving} and apply different prompt for each iteration. We trained with 300 iterations for our experiments. With a learning rate of 1e-4 and batch size of 1.

\noindent\textbf{LODA.} We set LODA step $N$ to 10. with percent hyperparameter $\gamma$ 0.9 and ReLU threshold $\tau$ to 1.0. We set each strength hyperparameter $p,m$ to +5 and -1e8. Update rate $\eta_t$ was scheduled linearly with $40 - 20 \cdot \frac{t}{T}$, where $T$ denotes the total steps.

\begin{table}[t]
\centering
\setlength{\tabcolsep}{8pt} 
\renewcommand{\arraystretch}{1.0} 
\scalebox{0.8}{ 
\begin{tabular}{cccccc}
\toprule
\textbf{\# of Concepts} & \textbf{Method} & \textbf{Capacity} & \textbf{Times} \\
\midrule
\multirow{6}{*}{\textbf{Single concept}} 
& Textual Inversion~\cite{gal2022textualinversion} & 4.2KB & {$\sim$~2h} \\
& DreamBooth~\cite{ruiz2023dreambooth} & 5.2GB  & {$\sim$~7m}\\
& Custom Diffusion~\cite{kumari2023custom}  & 97.5MB & {$\sim$~12m}\\
& Cones 2~\cite{liu2023cones2} & 4.2KB & $\sim$~35m \\
& EDLoRA~\cite{gu2024mixofshow} & 6.6 MB & $\sim$~28m \\
& \textbf{ConceptSplit (Ours)} & 7.4MB & \textbf{$\sim$~3m} \\
\midrule
\multirow{6}{*}{\textbf{Two concepts}} 
& Textual Inversion~\cite{gal2022textualinversion} & 8.4KB & {$\sim$~4h} \\
& DreamBooth~\cite{ruiz2023dreambooth} & 5.2GB  & {$\sim$~14m} \\
& Custom Diffusion~\cite{kumari2023custom} & 97.5MB & {$\sim$~24m}\\
& Cones 2~\cite{liu2023cones2} & 8.4KB  & $\sim$~70m \\
& EDLoRA~\cite{gu2024mixofshow} & 13.2MB & $\sim$~56m \\
& \textbf{ConceptSplit (Ours)} & 14.8MB  & \textbf{$\sim$~6m} \\
\midrule
\multirow{6}{*}{\textbf{Three concepts}} 
& Textual Inversion~\cite{gal2022textualinversion} & 12.6KB & {$\sim$~6h} \\
& DreamBooth~\cite{ruiz2023dreambooth} & 5.2GB & {$\sim$~21m} \\
& Custom Diffusion~\cite{kumari2023custom} & 97.5MB  & {$\sim$~36m}\\
& Cones 2~\cite{liu2023cones2} & 12.6KB  & $\sim$~105m \\
& EDLoRA~\cite{gu2024mixofshow} & 19.8MB & $\sim$~84m \\
& \textbf{ConceptSplit (Ours)} & 22.2MB  & \textbf{$\sim$~9m} \\
\midrule
\multirow{6}{*}{\textbf{Four concepts}} 
& Textual Inversion~\cite{gal2022textualinversion} & 16.8KB & {$\sim$~8h} \\
& DreamBooth~\cite{ruiz2023dreambooth} & 5.2GB  & {$\sim$~28m} \\
& Custom Diffusion~\cite{kumari2023custom} & 97.5MB  & {$\sim$~48m}\\
& Cones 2~\cite{liu2023cones2} & 16.8KB  & $\sim$~150m \\
& EDLoRA~\cite{gu2024mixofshow} & 19.8MB & $\sim$~112m \\
& \textbf{ConceptSplit (Ours)} & 29.6MB  & \textbf{$\sim$~12m} \\
\bottomrule
\end{tabular}
}

\caption{\textbf{Comparison of model capacity and training time across different personalization methods on Stable Diffusion v2.1.} This table presents a comparison of storage size (capacity) and training time for various personalization techniques, including Textual Inversion~\cite{gal2022textualinversion}, DreamBooth~\cite{ruiz2023dreambooth}, Custom Diffusion~\cite{kumari2023custom}, Cones 2~\cite{liu2023cones2}, EDLoRA~\cite{gu2024mixofshow}, and ConceptSplit (ours). ConceptSplit consistently demonstrates lower capacity requirements and faster training times.}
\label{tab:capacity_complexity_parameters}
\end{table}

\section{Analysis of Attention Entropy}
As shown in Figure 3 in the main paper, We found out that such key-modifying methods show disrupted attention, we first extract the attention map from U-net while forwarding, using the attention store class and storing every attention map for steps. After inference is ended, we aggregate these attention maps which have a resolution of 24 in every layer in U-Net. We extract the attention map from 24, as it is known to have the most semantic information \cite{rombach2022high}. Then averaged them and applied softmax to make them probability distribution. Then we calculated Entropy $\mathcal{H} = -\sum_{m,n} \hat{\mathbf{A}}(m,n) \log \hat{\mathbf{A}}_{(m,n)}$, where $\hat{A}\in\mathbb{R}^{24\times24}$ denotes aggregated, and normalized attention map. We calculated the average change of this entropy, which shows a detailed slope on \ref{tab:entropy_change}. We found out that when we modify these keys, the model gets confused, and through attention map seems to be noisy and disrupted as shown in Figure 4 in the main paper.
\label{sec:entropyanalysis}

\section{Qualitative results of ToVA Ablation.}
\label{sec:ToVAvisual}
We show our Qualitative results of ToVA ablation in Figure \ref{fig:tovtok}. These results show that modifying the key directly, results in degraded images.

\section{Algorithm of LODA}
We show an algorithm of LODA on Algorithm~\ref{alg:loda}, which we discussed in Section 3.3 in the main paper.

\section{Qualitative results of single Object}
\begin{figure*}[t]
    \centering
    \includegraphics[width=1\linewidth]{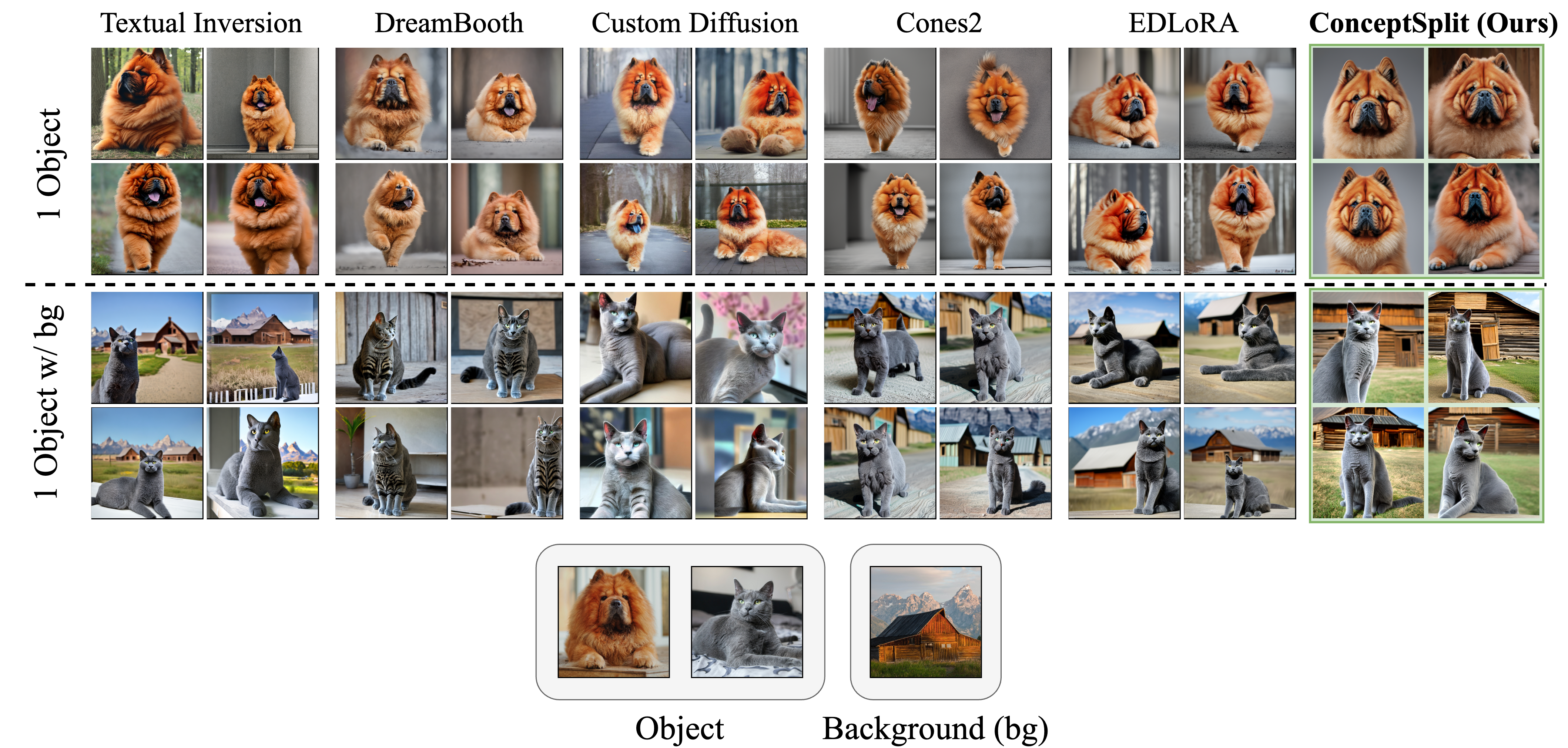}
    \caption{\textbf{Qualitative comparison in single-object scenarios on Stable Diffusion 2.1.} In single-object scenarios, our approach ensures that the background is appropriately generated alongside the target concept, maintaining contextual integrity.}
    \label{fig:qualitative_single_object}
\end{figure*}

We show qualitative results of single Object personalization in Figure \ref{fig:qualitative_single_object}. Our methods show comparable results to existing methods.

\section{LODA to Stable Diffusion}
\label{sec:lodatosd}
In Figure~\ref{fig:aaevsloda}, we compare the results of applying our LODA to the pre-trained Stable Diffusion~\cite{rombach2022high} model without personalization against Attend-and-Excite~\cite{chefer2023attendandexcite}. Attend-and-Excite focuses on increasing the maximum attention values, which often leads to significant concept mixing. In contrast, LODA  actively relocates and separates objects, enabling Stable Diffusion to effectively distinguish between them. This demonstrates that LODA is not effective in scenarios of personalization, rather it can boost the performance of the pre-trained Stable Diffusion model.

\section{Ours on SDXL}
We also implement our method to SDXL\cite{podell2023sdxl}, showing feasibility for both vanilla and personalized settings as shown in figure \ref{SDXL}.

\section{Hyperparameter Ablations}

\subsection{The Effect of the Percentile $\boldsymbol{\gamma}$}
\begin{figure*}
    \centering
    \includegraphics[width=0.9\linewidth]{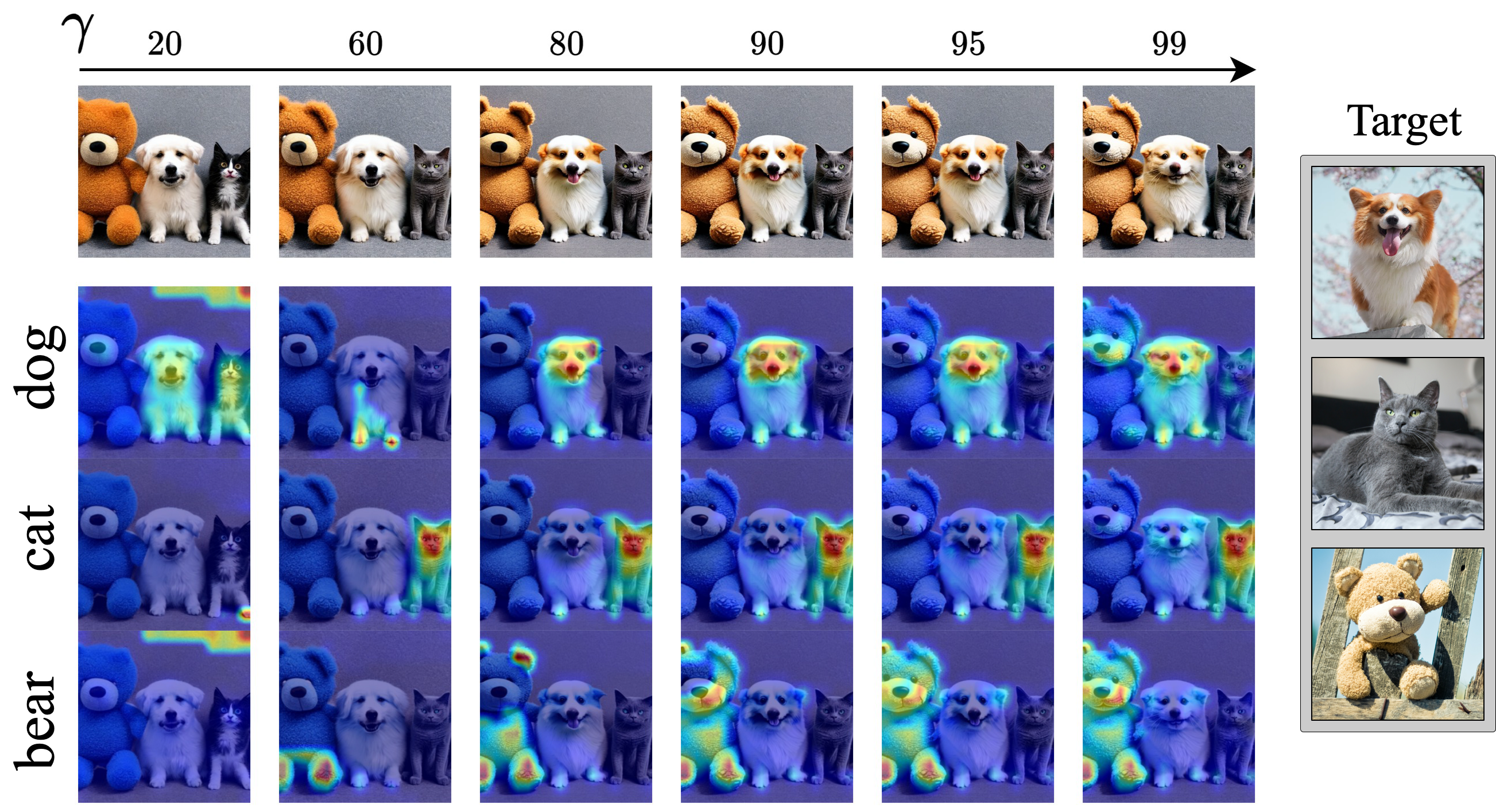}
    \caption{Effect of hyperparameters $p$ and $m$, which respectively strengthen and weaken the attention scores of each token.}
    \label{fig:gamma}
\end{figure*}

In Figure~\ref{fig:gamma}, we illustrate the visual effects of varying the percentile hyperparameter $\gamma$. 
A low $\gamma$ value means we consider broader attention regions for each token.
When applied to Attention Fixing Guidance, this broad consideration leads to excessive removal of overlapping areas, hindering successful personalization.
As $\gamma$ increases to around 80 or higher, proper fixation of attention occurs.
However, setting $\gamma$ too high, such as at 99, results in only very localized regions being fixed, failing to adequately suppress the influence of the ``cat" token. 
Consequently, this leads to images where, for example, a dog has cat whiskers, indicating that the ``cat" token's influence was not sufficiently reduced.

\subsection{The Effect of $\boldsymbol{p},\boldsymbol{m}$}
\begin{figure*}[t]
    \centering
    \includegraphics[width=0.9\linewidth]{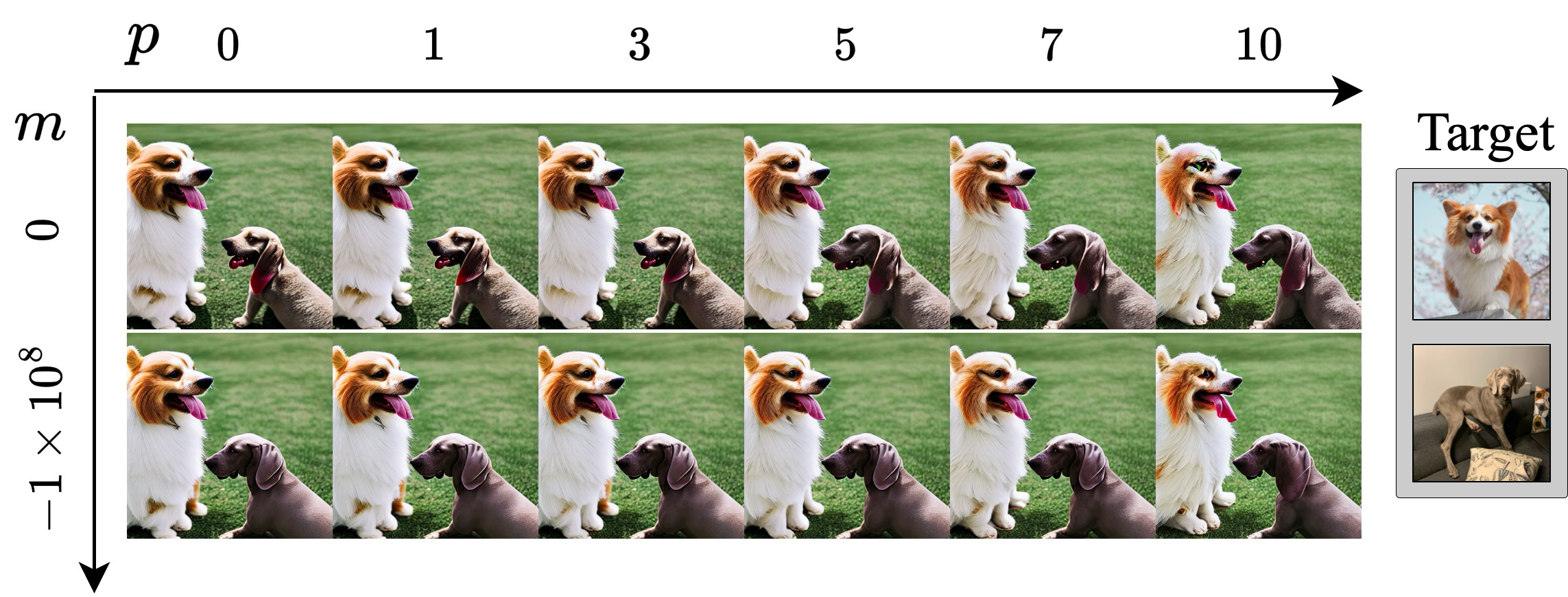}
    \caption{Effect of hyperparameters $p$ and $m$, which respectively strengthen and weaken the attention scores of each token.}
    \label{fig:pm}
\end{figure*}

\begin{figure*}[t]
    \centering
    \includegraphics[width=0.9\linewidth]{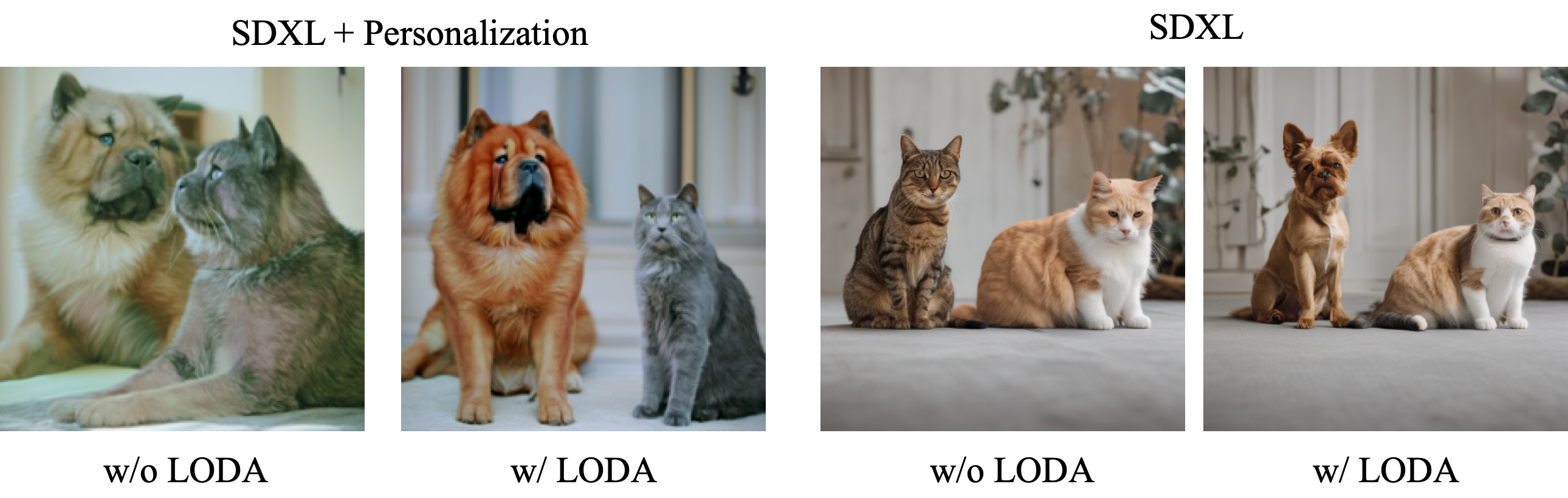}
    \caption{\textbf{Application of our method to SDXL.} Our approach is implemented on SDXL, demonstrating feasibility in both vanilla and personalized settings.}
    \label{SDXL}
\end{figure*}

\begin{figure*}[h]
    \centering
    \includegraphics[width=0.9\linewidth]{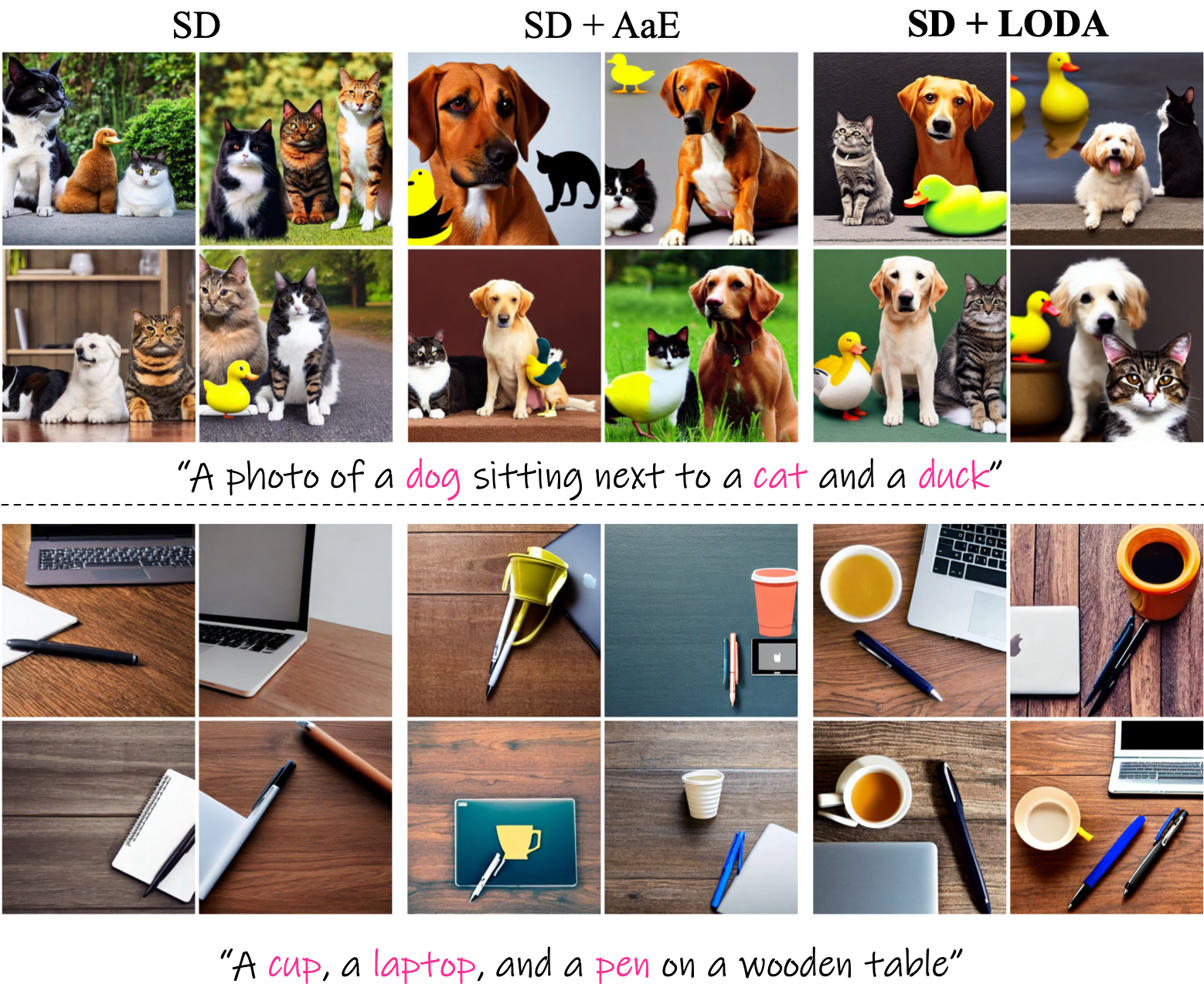}
    \caption{\textbf{Comparison of Attention-and-Excite (AaE) \cite{chefer2023attendandexcite} and LODA on Stable Diffusion 1.5.} This figure illustrates the differences in concept preservation and controllability between AaE and our proposed method, LODA. While AaE focuses on attention refinement to better represent multiple objects, LODA further enhances concept disentanglement, reducing interference between personalized concepts.}
    \label{fig:aaevsloda}
\end{figure*}

\begin{figure*}
    \centering
    \includegraphics[width=1\linewidth]{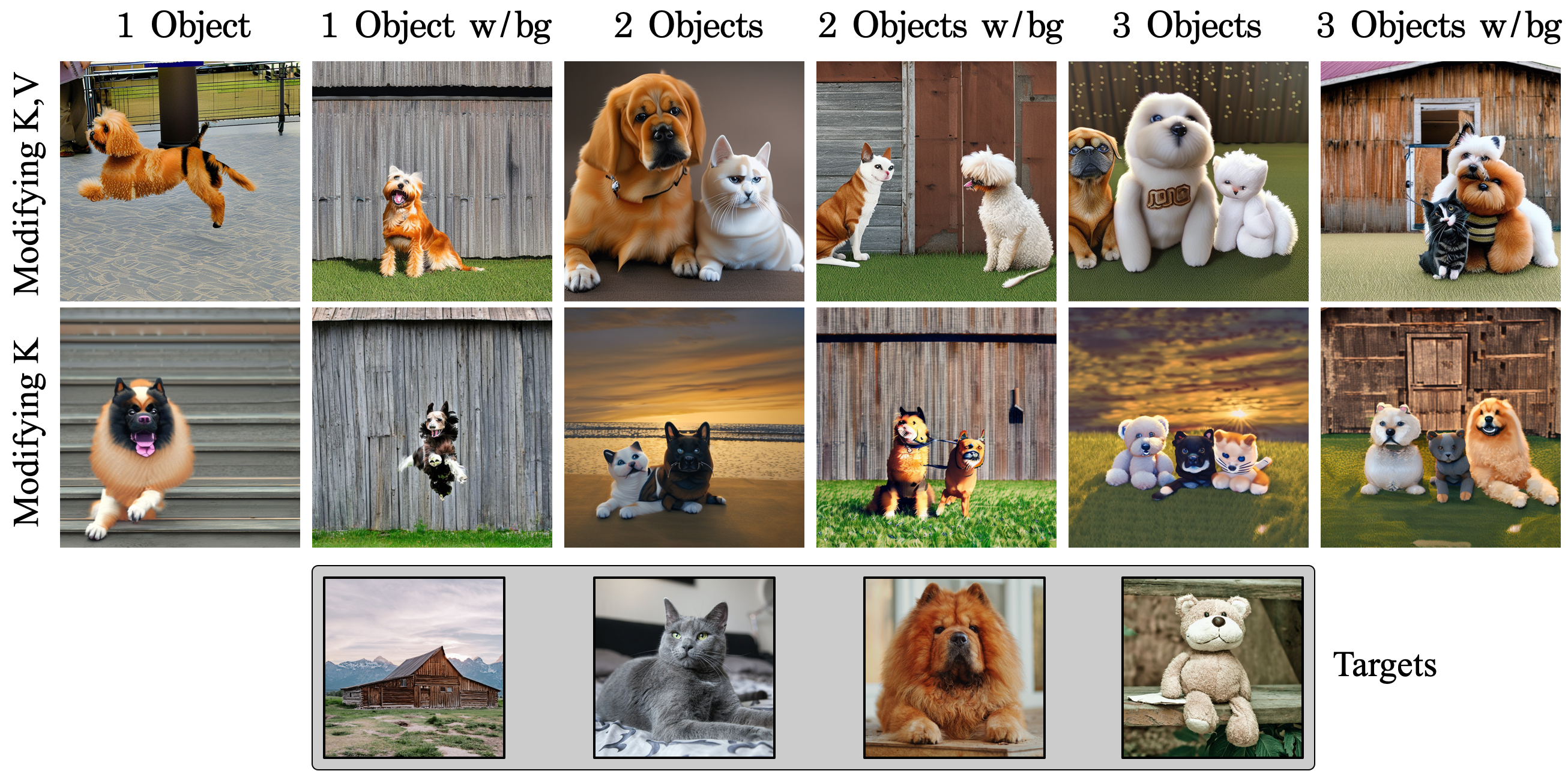}
    \caption{\textbf{Qualitative results of ToVA ablation.}}
    \label{fig:tovtok}
\end{figure*}

In Figure~\ref{fig:pm}, we present how the images change based on the parameters $p$ and $m$, which are used to strengthen or weaken attention, respectively. 
Adjusting the value of $p$ slightly enhances individual features but does not produce significant overall differences in the generated images.
However, applying the parameter $m$ has a substantial impact; when $m$ is applied, the influence of each token on other tokens diminishes, causing the learned concepts to appear more distinctly. 
In contrast, without applying $m$, the resulting images exhibit a mixed form due to overlapping token influences.



\section*{Acknowledgement}
This work was supported by Institute of Information \& communications Technology Planning \& Evaluation (IITP) grant funded by the Korea government (MSIT) (No. RS-2019-II190079, Artificial Intelligence Graduate School Program (Korea University), No. RS-2024-00457882, AI Research Hub Project, and Development of Semi-Supervised Learning Language Intelligence Technology and Korean Tutoring Service for Foreigners, under Grant 2019-0-00004).

\clearpage
{
    \small
    \bibliographystyle{ieeenat_fullname}
    \bibliography{main}
}


\end{document}